\documentclass[11pt]{article}
\usepackage{eacl2017}
\usepackage{times}
\usepackage{url}
\usepackage{latexsym}
\usepackage{amsmath}
\usepackage{graphicx}
\graphicspath{ {images/} }
\usepackage{color}

\eaclfinalcopy 
\def\eaclpaperid{***} 

\def\Tref#1{Table~\ref{#1}}
\def\Fref#1{Figure~\ref{#1}}

\def\Sref#1{Section~\ref{#1}}

\def\perscite#1{\newcite{#1}}   
\def\parcite#1{\cite{#1}}    

\def\footurl#1{\footnote{\url{#1}}}
\def\equo#1{``#1''}

\title{LanideNN: Multilingual Language Identification on Character Window}

\author{Tom Kocmi \and Ond{\v r}ej Bojar\\
  Charles University, Faculty of Mathematics and Physics \\
  Institute of Formal and Applied Linguistics \\
  {\tt \{kocmi,bojar\}@ufal.mff.cuni.cz} \\}

\date{}

\begin{document}
\maketitle
\begin{abstract}
In language identification, a common first step in natural language processing,  we want to automatically determine the language of
some input text. Monolingual language identification assumes that the given
document is written in one language. In multilingual language identification,
the document is usually in two or three languages and we just want their names. We aim one step further and
propose a method for textual language identification where languages
can change arbitrarily and the goal is to identify the spans of each of the
languages.

Our method is based on 
Bidirectional Recurrent Neural Networks and it performs well in monolingual and
multilingual language identification tasks on six datasets covering 131
languages.
The method keeps the accuracy
also for short documents and across domains, so it is
ideal for off-the-shelf use without preparation of training data.
\end{abstract}

\section{Introduction}

The World Wide Web is an ever growing source of textual data, especially data
generated by web users. As more 
people get access to the web, more languages and dialects start to appear and need to be processed.
In order to be able to use such data for further natural language processing
(NLP) tasks, we need to know in which languages they were written. 
Language identification is thus a key component for both building various NLP
resources from the web and also for running many web services.

Techniques of language identification can rely on handcrafted rules, usually of
high precision but low coverage, or data-driven methods that learn to identify
languages based on sample texts of sufficient quantity.

In this paper, we present a data-driven method for language identification based
on bidirectional recurrent neural networks called \emph{LanideNN} (language
identification by neural networks, NN).
The model is trained on character sliding window of input texts with the goal of assigning a language to
each character.
%
We show that the method is applicable for a large number of languages and across text domains without any
adaptation and that it performs well in monolingual (one language per document)
as well as multilingual (a few languages per document)
language identification tasks. Also, the performance does not drop with shorter
texts.

The paper is structured as follows.
In \Sref{related}, we briefly review current approaches to language
identification. \Sref{method} introduces our method, including the technical
details of the neural network architecture. For the training of our model, we
collect and manually clean a new dataset, as described in \Sref{data}. The
model is evaluated on standard test sets for monolingual (\Sref{mono}) as well
as multilingual (\Sref{multi}) language identification. \Sref{partitioning}
illustrates the behavior of our method in the motivating setting: identifying
languages in short texts. We conclude and summarize our plans in
\Sref{conclusion}.

\section{Related Work}
\label{related}

Of the many possible approaches to language identification
\perscite{Hughes06reconsideringlanguage}, character $n$-gram
statistics are among the most popular ones.
\perscite{cavnar1994} were probably the first; they used the 300 most frequent
character $n$-grams (with $n$ ranging from 1 to 5, as is also typically used in
other works).
All the $n$-gram-based approaches differ primarily in the calculation of the distance
between the $n$-gram profile of the training and test text
\parcite{Selamat2011,Yang2010}, or by using additional features on top of the
$n$-gram profiles \parcite{Padma2009,Carter2013}.
One of the fairly robust definitions of the distance (or similarity) was
proposed by
\perscite{Choong2009}
who simply check the proportion of $n$-gram types seen in
the tested document of the most frequent
$n$-gram types extracted
from
training documents for each language. The highest-scoring language is then
returned.

\perscite{Hughes06reconsideringlanguage} mention a number of freely available
tools at that time. Since then, one aspect of the tools became also important: the number of languages
covered.

The language identification tool
CLD2\footurl{https://github.com/CLD2Owners/cld2} by Google detects 80 languages
and uses a Naive Bayes
classifier, treating specifically unambiguous scripts such as Greek and using
either character unigrams (Han and similar scripts) or fourgrams.

Another popular tool is Langid.py by \perscite{Lui2012}, covering 97 languages
out of the box. Langid.py relies on Naive Bayes
classifier with a multinominal event
model and mixture of byte $n$-grams for training. The tool includes tokenization
and fast feature extraction using Aho-Corasick string matching.

To our knowledge, and also according to the survey by \perscite{survey2014},
neural networks have not been used often for language identification so far. One
exception is \perscite{Dubaee2010}, who combine a feed-forward network
classifier with wavelet transforms of feature vectors to identify English and
Arabic from the Unicode representation of words, sentences or whole documents.
The benefit of NN in this setting is not very clear to us because English and
Arabic can be distinguished by the script.
During writing of
this paper, we have found a new pre-print paper
\parcite{jaech2016hierarchical} which handles language identification with NN.
Specifically, they employ Convolutional Neural Networks followed by Recurrent Neural
Networks. Their approach labels text on the word level, which is problematic in 
languages without clear word delimiters. In comparison with our model,
they need to pre-process the data and break long words into smaller chunks,
whereas we simply use text without any preprocessing.

In practice, several tools are often used at once, with some form of majority voting.
For example, Twitter internal language detector uses their in-house tool along with
CLD2 and Langid.py, and this triple agreement is reported to make less than
1\,\% of
errors.\footurl{https://blog.twitter.com/2015/evaluating-language-identification-performance}

Multilingual language identification, i.e. identification of the set of
languages used in a document, is a less common task, explored e.g. by
\perscite{Lui2014} who use a generative mixture model on multilingual documents
and establish the relative proportion of languages used. Character $n$-grams
again serve as features, selected by information gain.

\perscite{solorio-EtAl:2014:CodeSwitch} organized a shared task in language
identification at the word level. This matches our aim, but the task included only four language
pairs and more importantly, the dataset was collected from Twitter and for
copyright reasons it is not available any more.

\section{Proposed Method}
\label{method}

The method we propose is designed for short text without relying on document
boundaries. Obviously, if documents are known and if they can be assumed to be
monolingual, this additional knowledge should not be neglected. For the long
term, we however aim
at a streamlined processing of noisy data genuinely appearing in multilingual
environments. For instance, our method could support the study of switching of
languages (\equo{code switching}) in
e-mails or other forms of conversation, or to analyse various online media
such as Twitter, see e.g. \perscite{montes-alcala:blogging:2007} or
\perscite{solorio-EtAl:2014:CodeSwitch}.

Our model takes source letters as input and
provides a language label for each of them. Whenever we need to recognize the language
of a document, we take the language assigned by our model to the majority of
letters.

The goal of attributing a language tag to the smallest text units is one of the
reasons why we decided to use neural networks and designed the model to provide
a prediction at every time step without much overhead.

In the rest of this section, we explain the architecture and training methods of the
model.

\subsection{Bidirectional Recurrent Neural Networks}

\begin{figure*}[t]
\begin{center}
\includegraphics[width=0.8\textwidth]{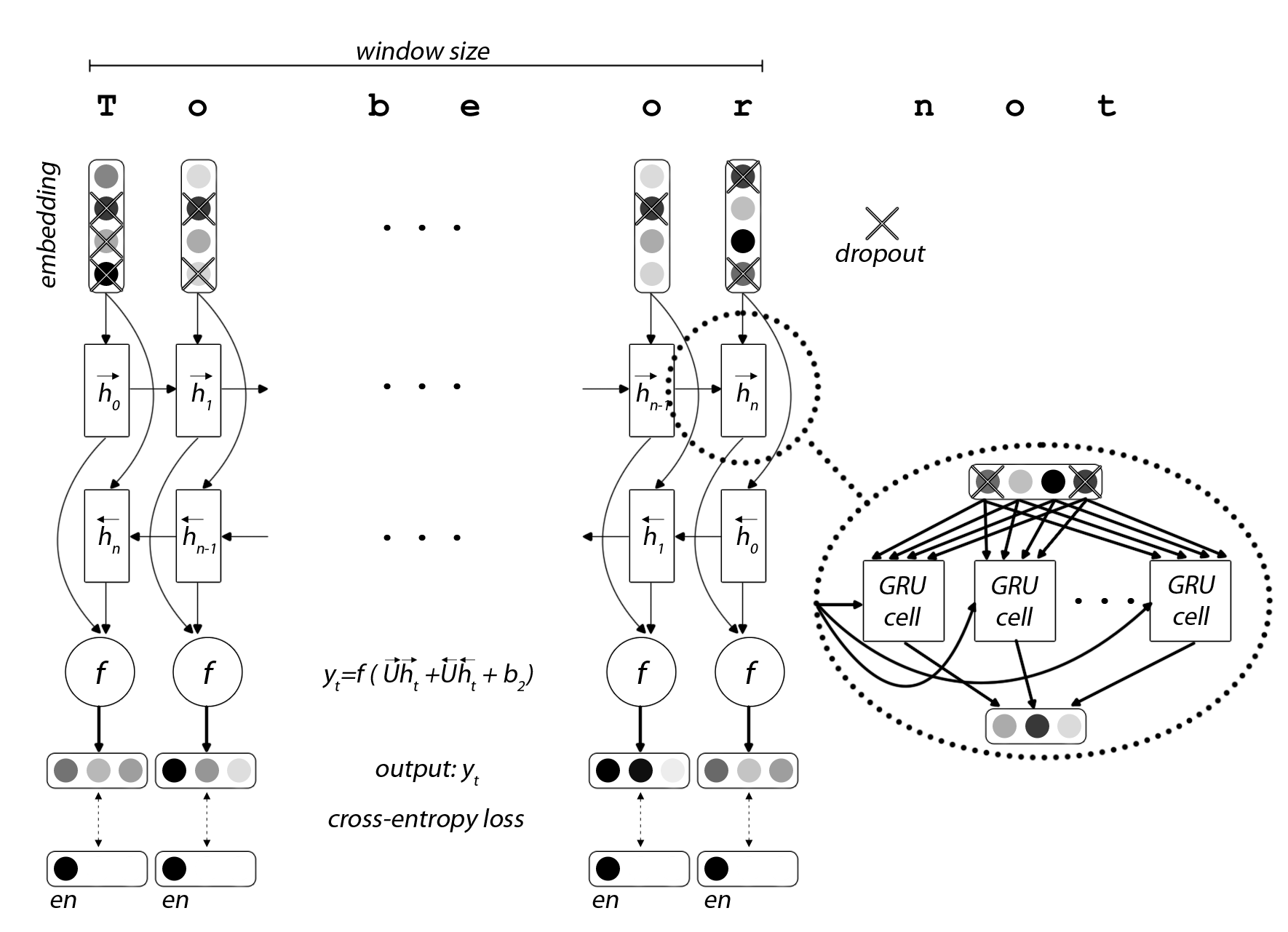}
\end{center}
\caption{\label{model}Illustration of our model LanideNN.
}
\end{figure*}

A recurrent neural network \emph{RNN} \parcite{Elman1990} is a variant of
neural networks with recurrent connections in time.
In principle, the history information available to an RNN is not limited (subject
to a processing window, if used), so
the network can condition its output on features from a long distance.
The LSTM,
one of the variants of RNN, makes it particularly suitable for sequential prediction
tasks with
arbitrary time dependencies, as shown by \perscite{Hochreiter1997}.


In this work, we use the Elman-type network, where the hidden layer $h_t$ at a
time step $t$ is computed based on the current input layer $x_t$ and the
previous state of the hidden layer $h_{t-1}$. The output $y_t$ is then derived
from the $h_t$ by applying the softmax function $f$. More formally:

\vspace{-0.8em}
\begin{align}
h_t &= \tanh \left( Wx_t + Vh_{t-1} + b_1 \right) \\
y_t &= f \left( Uh_t + b_2 \right)
\end{align}

\noindent where $U$, $V$ and $W$ are connection weights to be computed in training time and bias vectors $b_1$ and $b_2$.

With the above definition, the RNN has access only to information preceding the
current position in the text. In our setting, the rest of the text (in a
fixed-size window) is
available, so we want to allow the model to use also future information, i.e.
letters following the currently examined one.
We therefore define a second RNN which reads the input from the
end to the beginning, changing the definition to:

\vspace{-0.8em}
\begin{align}
\overrightarrow{h}_t &= \tanh \left( \overrightarrow{W}x_t + \overrightarrow{V}\overrightarrow{h}_{t-1} + \overrightarrow{b}_1 \right)
\\
\overleftarrow{h}_t &= \tanh \left( \overleftarrow{W}x_t + \overleftarrow{V}\overleftarrow{h}_{t+1} + \overleftarrow{b}_1 \right)
\\
y_t &= f \left( \overrightarrow{U}\overrightarrow{h}_t +\overleftarrow{U}\overleftarrow{h}_t + b_2 \right)
\end{align}

\noindent where the left and right arrows indicate the direction of network. 


The simple unit with only tanh non-linearity is difficult to train and therefore
we have selected the Gated Recurrent Unit (GRU),
recently proposed by \perscite{Cho2014}, as a replacement. We also considered Long Short-Term
Memory cells (LSTM) but they achieved slightly worse results in our setting.
This changes equations (1), (3) and (4). The proper equations for the GRU can be
found in \perscite{Cho2014}.

The model outputs a probability distribution over all language tags. In order to
determine the language of a character, we take the tag with the maximum value.

The complete model is sketched in \Fref{model}.


\subsection{Training, Embeddings and Dropout}

We train the model using the first-order stochastic gradient descent method Adam \parcite{Kingma2014}. Our training
criterion is the cross-entropy loss function\footnote{We set the learning rate to
0.0001 and train with the batch size of 64 windows.}. 

We represent each Unicode character using an $e$-dimensional real valued vector,
analogously to word embeddings of \perscite{FromScratch}. The character
embeddings are initialized randomly and are trained together with the rest of the network.



To prevent overfitting, we use dropout \parcite{dropout2014} during model training on the character embedding layer\footnote{We set the dropout to the probability of 0.5 as customary.}.
The key idea is to randomly drop (avoid updating of) connections. This prevents neurons from
co-adapting too much, i.e. starting to depend on outputs of other neurons too
much, which is a typical symptom of overfitting to training data.

\subsection{Model Design}

Our model operates on a window of 200 characters of input text, i.e. individual letters, encoded in Unicode.
Each character corresponds to one time step of the BiRNN in the respective
direction, see \Fref{model}.
The model classifies each character separately, but quickly
learns to classify neighbouring characters with the same label.

For documents longer than the window size, we simply move to the next
window without any overlap. The last window (or the only window if the
document were too short) is filled with a padding character, so the network
always works on windows of the same size.

We set $e$, the size of the embedding layer, to 200. The BiRNN uses a single hidden
layer of 500 GRU cells for each direction.

The main model was trained for over 530,000 steps (each step is the processing
of one batch of inputs) on a single core of the GeForce GTX Titan Z GPU. The training
took around 5 days. The stopping criterion for the training was the error on a 
development set.

\section{Training Data}
\label{data}

Our goal is to develop an off-the-shelf language recognizer, with no need for
retraining by the user and covering
as many languages as possible. Finding suitable training data is thus an
important part of the endeavour.

We start from Wikipedia, as crawled and converted to a large multilingual corpus
W2C by \perscite{Majlis2012}. W2C contains 106 languages but we had to exclude a
few of them\footnote{Specifically, HAT, IDO, MGL, MRI,
VOL, as identified by ISO language codes.} because they contained too little non-repeating text.

\begin{figure}[t]
\begin{center}
\framebox{
\begin{minipage}{.95\columnwidth}
\footnotesize
afr, amh, ara, arg, asm, ast, aze, bak, bcl, bel, ben, ber, bpy, bre, bul, cat,
ceb, ces, che, chv, cos, cym, dan, deu, div, ekk, ell, eng, est, eus, fas, fin,
fra, fry, gla, gle, glg, gom, gsw, guj, hat, heb, hif, hin, hrv, hsb, hun, hye,
ido, ilo, ina, ind, isl, ita, jav, jpn, kal, kan, kas, kat, kaz, kir, kor, kur,
lat, lav, lim, lit, ltz, lug, lus, mal, mar, min, mkd, mlg, mlt, mon, mri, msa,
nds, nep, new, nld, nno, nor, nso, oci, ori, oss, pam, pan, pms, pnb, pol, por,
pus, roh, ron, rus, sah, scn, sin, slk, slv, sna, som, spa, sqi, srp, sun, swa,
swe, tam, tat, tel, tgk, tgl, tha, tur, uig, ukr, urd, uzb, vec, vie, vol, wln,
yid, zho, zul
\end{minipage}
}
\end{center}
\caption{The 131 languages (and HTML) recognized by our system.}
\label{langs}
\end{figure}

\begin{table*}[t]
\begin{center}
\small
\begin{tabular}{cccccc}
Test Set & Documents & Languages & Encoding & Document Length (bytes) & Avg. \# characters \\
\hline
EuroGov & 1500 & 10 & 1 & 17460.5 $\pm$ 39353.4 & 17037.3\\
TCL & 3174 & 60 & 12 & 2623.2 $\pm$ 3751.9 & 1686.1\\
Wikipedia & 4963 & 67 & 1 & 1480.8 $\pm$ 4063.9 & 1314.2\\
\end{tabular}
\end{center}
\caption{Summary of testsets for monolingual language identification.}
\label{testsets}
\end{table*}

We then focussed on finding corpora with 
at least some languages not covered
by the already collected data. Those corpora were added whole, including
languages that we already had, to improve the variety of our collection.
We made use of the following ones:

\begin{description}
  \item[Tatoeba\footnotemark] \footnotetext{\url{http://tatoeba.org/}}is a
collection of simple sentences
for language learners. Tatoeba contains sentences in 307 languages, but for most
languages it has only a few hundred sentences.

\item[Leipzig corpora collection] \parcite{Quasthoff2006} covers 220 languages
with newspaper text and other various texts collected from the web.

\item[EMILLE] \parcite{Baker2002} contains texts in 14 Indian languages
and English.

\item[Haitian Creole training data] \parcite{callisonburch-EtAl:2011:WMT}
were prepared by the organizers of WMT11 shared task on machine translation of
SMS messages sent to an emergency hotline in the aftermath of the 2010 Haitian
earthquake. We used only the official documents from the training data, not the
actual SMS messages because they contained a lot of noise.

\end{description}

Additionally, we wanted our tool to
distinguish HTML tags in the data, since they are the most frequent markup that
needs to be separated from the processed data. Therefore, we have downloaded
several Github projects in HTML and collected all strings enclosed with angle
brackets, as a rather permissive approximation of HTML tags. We have dropped
tags which were too long and we put each tag on a separate line. We
have not deduplicated them for the training set. 

The cleanup of the collected data was mostly manual. We deduplicated each of the
sources by dropping identical lines, regardless of what lines correspond to in the
individual sources (words, phrases, sentences or even paragraphs). 
We inspected data files for individual languages and removed lines containing
English for languages not using Latin script. We also removed Cyrilic characters
from a few languages that should not contain them. This was done mostly in W2C corpora. 


For the final dataset, we mixed
all sources for a given language at the line
level, keeping only languages with more than 500k characters in total.
Since the resources for some languages were huge, we
decided to set an upper bound on the number of characters per
language. In order to roughly reflect the distribution of languages in the
world, we
divided languages into three groups based on the number speakers of the
language according to Wikipedia.
The first group were languages
with more than 75M speakers, the second with more than 10M speakers and the third
group contained the rest. For the first group, we allowed at most 10M
characters in the training set, the second group was capped at 5M characters and the third group was allowed
only 1M characters per language at most.\footnote{Higher-quality sources such as
Tatoeba are generally smaller and since we mixed the sources by interleaving
their lines, these smaller sources were likely included in full.}

In total, our final training set includes 131 + 1 (HTML) languages, see
\Fref{langs}.

\label{our-test-set}
We divide the corpus into non-overlapping
training, development and test sections. We released the test
set \footnote{https://ufal.mff.cuni.cz/tom-kocmi/lanidenn} but the training part cannot be publicly released because of the restrictive permissions
of some of the sources used.  The test section is
limited to short text. It contains 100 lines for each of the 131 languages (HTML
is not included), with
the average line length of 142.3 characters.

Each line of the dataset starts with an ISO-3 label of the language presented on
that line. All lines were shuffled.

For training and testing, the language labels as well as all line breaks must be ignored, otherwise the model could learn to
set language boundaries at the new line character. After dropping all line
breaks, we obtain a multilingual text.

This way, we simulate a multilingual text and our algorithm has to learn to
identify language boundaries without relying on any particular symbol. We are aware of
the fact that the original segmentation of the corpora affects where these
language switches are expected, and this will mostly correspond to sentence
boundaries.

\begin{table*}[t]
\begin{center}
\small
\begin{tabular}{l|c|c|ccc}
System & Trained on & Supported languages & EuroGov & TCL & Wikipedia \\
\hline
LangDetect* & Wikipedia & 53 & \textbf{.992}\rlap{$^\text{\ref{langdetect-eurogov}}$} & .818 & .867\\
TextCat* & TextCat Dataset & 75 & .941 & .605 & .706\\
CLD* & unknown & 64 & .983 & .732 & .831\\
Langid.py* & \perscite{Dataset2011} & 97 & .987 & .904 & \textbf{.913} \\
\hline
Langid.py & \perscite{Dataset2011} & 97& .987 & .931 & \textbf{.913}\\
CLD2 & unknown & 83& .979 & .837 & .854 \\
Our model & Our dataset & \textbf{136} &  .977 & \textbf{.954} &  .893\\
\end{tabular}
\end{center}
\caption{
Results of monolingual language identification on the \perscite{Baldwin2010Dataset} test set. Entries marked with \equo{*}
are accuracies reported by \perscite{Lui2012}, the rest are our measurements.
}
\label{langidCorpus}
\end{table*}

\section{Monolingual Language Identification}
\label{mono}

Most of related research is focused on monolingual language
identification, i.e. recognizing the single language of an input document.

We compare our method in this setting with several other 
algorithms on the dataset presented by
\perscite{Baldwin2010Dataset}. The dataset consists of 3 different test sets, each containing
a different number of languages, styles and document lengths collected from
different sources, see \Tref{testsets} for details:


\begin{description}
\item[EuroGov] contains texts in Western European
languages from European government resources.
\item[TCL] was extracted by the Thai Computational
Linguistics Laboratory in 2005
from online news sources and the test set also contains multiple file encodings.
Since our method assumes Unicode input, we converted TCL to Unicode encoding.
\item[Wikipedia] are texts collected from a
Wikipedia dump.
\end{description}

\Tref{langidCorpus} summarizes the accuracies of several algorithms on the three
test sets.
For some algorithms, we report
values as presented by \perscite{Lui2012} without
re-running\footnote{\label{langdetect-eurogov}We should mention that LangDetect used EuroGov as a validation set, so its score
on this test set is not reliable.}.
%
We re-ran the Langid.py as the best algorithm out of them, and got the
same results except for the TCL testset, where we got better results than reported
by \perscite{Lui2012}.
After a discussion with the authors, we believe the re-run benefited from the conversion
of all texts to Unicode.

We compare our method with two top language recognizers, Langid.py and CLD2. Our
model is trained on more languages and we do not restrict it to the languages
included in the test set, so we may be losing on detailed dialect labels.


Despite the considerably higher number of languages covered, our model performs
reasonably close to the competitors on EuroGov and Wikipedia and best on TCL.

\subsection{Short-Text Language Identification}

In order to 
demonstrate the ability of our method to identify language of very short texts
such as tweets, search queries or user messages, we wanted to use an existing
corpus, such as the one released by
Twitter.\footurl{https://blog.twitter.com/2015/evaluating-language-identification-performance}
Unfortunately, the corpus contains only references to the actual tweets and most
of them are no longer available.
We thus 
have to rely on our own test set, as described in \Sref{our-test-set}.

\begin{table}[t]
\begin{center}
\small
\begin{tabular}{l|c|c}
System & All languages & Common languages \\
\hline
Langid.py & .567 & .912 \\
CLD2 & .545 & .891 \\
Our model & \textbf{.950} & \textbf{.955}\\
\end{tabular}
\end{center}
\caption{Results on our test set for short texts. The first column shows accuracy over all 131 languages and the second column shows accuracy over languages that all systems have in common.}
\label{short-text-results}
\end{table}

\begin{table*}[t]
\begin{center}
\small
\begin{tabular}{l|c|cccccc}
System &
Training set & $P_M$ & $R_M$ & $F_M$ & $P_\mu$ & $R_\mu$ & $\mathbf{F_\mu}$ \\
\hline
\perscite{Baldwin2010} * & ALTW2010 & .497 & .467 & .464 & .833 & .826 & .829 \\
ALTW2010 winner * & ALTW2010 & .718 & .703 & .699 & .932 & .931 & .932 \\
SEGLANG * & ALTW2010 - mono & .801 & \textbf{.810} & .784 & .866 & .946 & .905 \\
LINGUINI * & ALTW2010 - mono & .616 & .535 & .513 & .713 & .688 & .700 \\
\perscite{Lui2014} * & ALTW2010 - mono & .753 & .771 & .748 & .945 & .922 & .933 \\
\hline
\perscite{Lui2014} our retrain & ALTW2010 - mono & .768 & .716 & .724 & \textbf{.968} & .896 & .931 \\
Our model & ALTW2010 - mono & \textbf{.819} & .764 & .779 & .966 & \textbf{.964} & \textbf{.965} \\
Our model & Our dataset & .709 & .714 & .695 & .941 & .941 & .941\\
\end{tabular}
\end{center}
\caption{\label{ALTW2010}Results of multilingual language identification on the
ALTW2010 test set. 
* As reported by \perscite{Lui2014}
}
\end{table*}

\begin{table*}[t]
\begin{center}
\small
\begin{tabular}{l|cccccc}
System & $P_M$ & $R_M$ & $F_M$ & $P_\mu$ & $R_\mu$ & $\mathbf{F_\mu}$ \\
\hline
SEGLANG * & .809 & \textbf{.975} & .875 & .771 & \textbf{.975} & .861 \\
LINGUINI * & .853 & .772 & .802 & .838 &.774 & .805 \\
\perscite{Lui2014} * & .962 & .954 & .957 & \textbf{.963} & .955 & .959 \\
\hline
\perscite{Lui2014} our retrain & .962 & .963 & .961 & \textbf{.963} & .964 & .963 \\
Our model trained on WikipediaMulti & .962 & .974 & \textbf{.966} & .954 & .974 & \textbf{.964}\\
Our model trained on our dataset & .774 & .778 & .774 & .949 & .972 & .961\\
Our model trained on our dataset, restricted & \textbf{.966} & .973 & \textbf{.966} & .956 & .973 & \textbf{.964}\\
\end{tabular}
\end{center}
\caption{Results of multilingual language identification on the WikipediaMulti
test set.
* As reported by \perscite{Lui2014}}
\label{wikimultiresults}
\end{table*}


Results on short texts are reported in \Tref{short-text-results}. The two other
systems, Langid.py and CLD2 cover fewer languages and they were trained on texts
unrelated to our collection of data. It is therefore not surprising that they
perform much worse when averaged over all languages.

For a fairer comparison, we report also accuracies on a restricted version of
the test set that included only languages supported by all the three tested
tools. Both our competitors are meant to be generally applicable, so they should
(and do) perform quite well. Our system nevertheless outperforms them, reaching
the
accuracy of 95.5. Arguably, we can be benefitting from having trained on
(different) texts from the same sources as this test set.

\Tref{confusion} lists the most frequent misclassifications of our model on our
test set (unordered language pairs) of the 13100 items in the test set. The most
common error is confusing Indonesian with Modern Standard Arabic, which
indicates some noise in our training data rather than difficulty of separating
these two languages. The following pairs are expected: Standard Estonian (ekk)
vs. Estonian (est, a macro language which includes Standard Estonian), Bashkir
vs. Tatar,  Croatian vs. Serbian,
Asturian vs. Spanish, \dots

\def\sep{$\leftrightarrow$}
\begin{table}
\begin{center}
\small
\begin{tabular}{lr|lr|lr}
ind\sep{}msa  & 64  & ekk\sep{}est  & 36  & bak\sep{}tat  & 28 \\
hrv\sep{}srp  & 17  & glg\sep{}por  & 17  & nno\sep{}nor  & 16 \\
ast\sep{}spa  & 15  & fas\sep{}pus  & 13  & ces\sep{}slk  & 13 \\
hrv\sep{}slv  & 10  & dan\sep{}nor  & 10  & nep\sep{}new  & 8 \\
aze\sep{}tur  & 7  & mar\sep{}new  & 6  & ceb\sep{}tgl  & 6 \\
cat\sep{}spa  & 6  & arg\sep{}spa  & 6  & fra\sep{}oci  & 5
\end{tabular}
\end{center}
\caption{Most frequent confusions on our test set.}
\label{confusion}
\end{table}

Finally, our model is trained to distinguish also HTML as one additional
language. We did not include HTML in our test corpus but to satisfy the
requests of one of our reviewers, we checked the performance on our development
corpus: only one Portuguese and one Yakut segment was classified as HTML and
none of the 100 HTML segments were misclassified.

\section{Multilingual Language Identification}
\label{multi}

In multilingual language identification, systems are expected to report the set
of languages used in each input document. The evaluation criterion is thus
macro-
(M) or micro- ($\mu$) averaged precision (P), recall (R) or F-measure
(F).\footnote{
Note that for comparability with results reported in other works, macro-averaged
F-score is calculated as average over individual F-scores instead of the
harmonic
mean of $P_M$ and $R_M$. $F_M$ can thus 
fall out of the range between  $P_M$ and $R_M$.
}

We evaluate our model on two existing test sets for multilingual identification,
ALTW2010 shared task
and WikipediaMulti.
%
We are mainly interested in the performance of our general model, trained on all
our training data, on these test sets. But since
both test sets come with training data, we also retrain our model to test its
in-domain performance.
We limit the training of these specific models to 140,000 training steps for ALTW2010 and
75,000 steps for WikiMulti, keeping other settings identical to the main model.
Each training step amounts to the processing of 64 batches
of 200 letters of input.
The number of steps for both tasks was established by testing the error on the
development parts of the datasets.

To interpret the character-level predictions by our model for multilingual
identification,
we used the ALTW2010
development data to empirically set the threshold: if a
language is predicted for more than 3\,\% of characters in the document, 
it is considered as one
of the languages of the document.

\begin{figure*}[t]
    \centering
    \includegraphics[width=\textwidth]{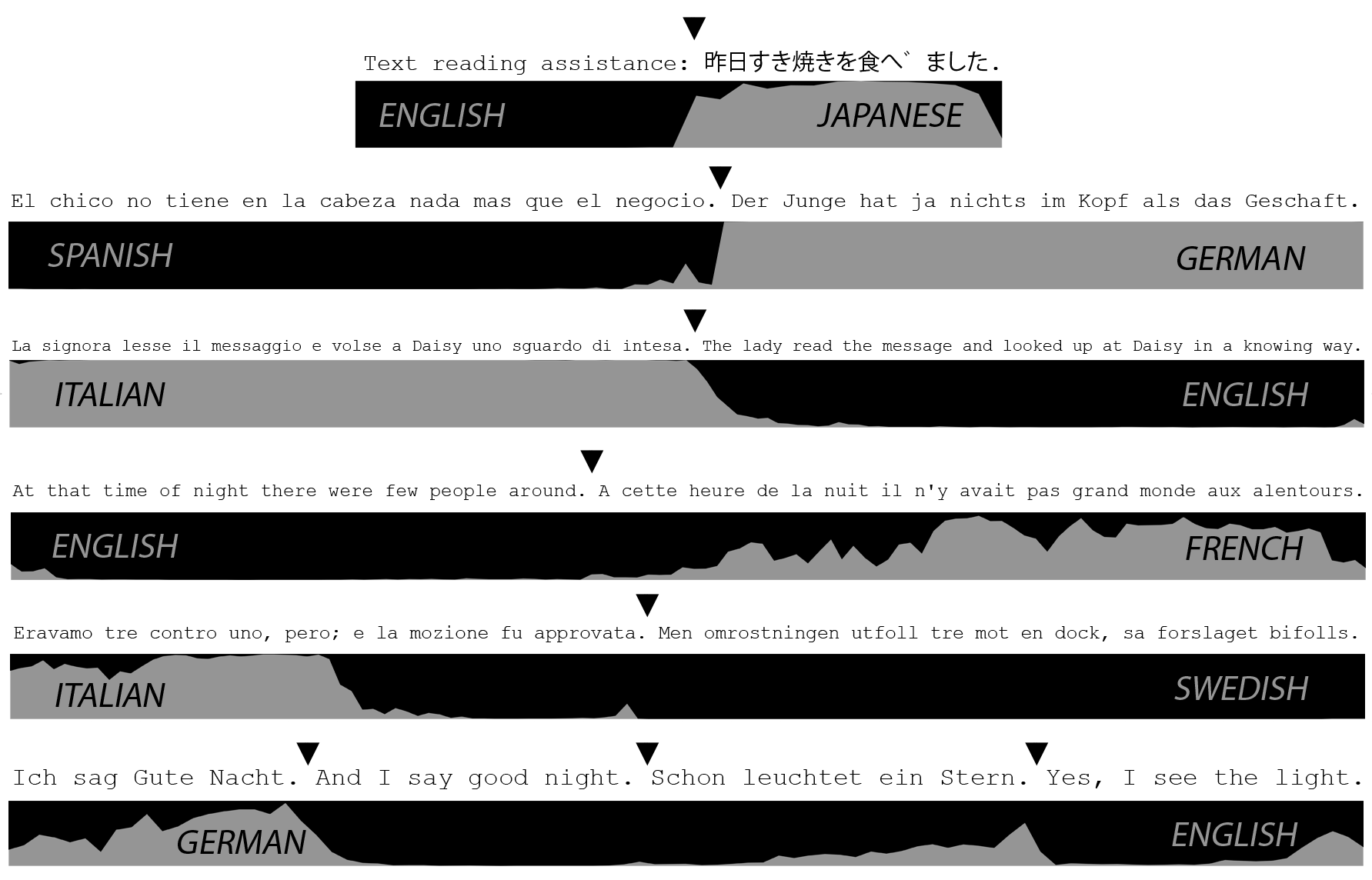}
    \caption{\label{crossing}Illustration of text partitioning. The black
triangles indicate true boundaries of languages. The black part shows
probability with which the language written in gray is detected and the gray
part shows complement for the  second language, since in
this setup we restricted our model to use only the two languages in question.
The misclassification of Italian and German as English in the last two examples
may reflect increased noise in our English training data.}
\end{figure*}

\subsection{ALTW 2010 Shared Task}

ALTW 2010 shared task \cite{Baldwin2010} provided 10000 bilingual documents
divided as follows: 8000 training, 1000 development and 1000 test documents.

The results on the 1000 test documents are in \Tref{ALTW2010}.
For algorithms SEGLANG and LINGUINI, we only reproduce the results reported by
\perscite{Lui2014}. We use the system by \perscite{Lui2014} as a proxy for the
comparison: we retrain their system and obtain results similar to those
reported by the original authors. The differences are probably due to the Gibbs
sampling used in their approach.



Some of the reported methods rely on the fact that the documents in the dataset
are bilingual. Other methods, including ours, simply break the bilingual
documents into the individual languages and train on this simplified training
set. We indicate this by stating \equo{ALTW2010 - mono} in \Tref{ALTW2010}.

The main criterion of the ALTW2010 shared task was to maximize the
micro-averaged F-score ($\mathbf{F_\mu}$). 
We see that our model trained on the ALTW2010 data outperforms all other models
in this criterion ($\mathbf{F_\mu}$ of .965) and so does our non-adapted
version, reaching $\mathbf{F_\mu}$ of .941.

\subsection{WikipediaMulti}

WikipediaMulti \parcite{Lui2014} is a dataset of artificially prepared
multilingual documents, mixed from monolingual Wikipedia articles from 44 languages. Each of
the artificial documents contains texts in $1\leq k \leq 5$ randomly selected
languages.
The average document length
is 5500 bytes. The training set consists of 5000 monolingual documents, the
development set consists of 5000 multilingual documents and test set consists of
1000 documents for each value of $k$. 

\Tref{wikimultiresults} shows that our model performs well, both when trained on the provided data
and when trained on our training corpus.
The model trained on our dataset performs slightly worse
in $\mathbf{F_\mu}$, but if we simply prevent it from predicting languages not present
in the test set, the score gets on par with the adapted version, see the line
labelled \equo{restricted} in \Tref{wikimultiresults}.

\section{Text Partitioning}
\label{partitioning}

\Fref{crossing} illustrates the behaviour of our model on text with mixed
languages. We have selected very short (50--130 characters) and challenging
segments where the languages mostly share the
same script. Finding the boundary between languages written in different scripts
is quite easy, as illustrated by the first example.

Only too late, we discovered that \perscite{king-abney:2013:NAACL-HLT} provide a test
set for word-level identification for 30 languages. We thus have to leave the
evaluation of our model on this dataset for future.


%


%
%
%
%

\section{Conclusion}
\label{conclusion}

We have developed a language identification algorithm based on bidirectional
recurrent neural networks.
The approach is designed for identifying
languages on a short texts, allowing to detect code switching including switches
to formal
markup languages like HTML.

We collected a dataset and trained our model to recognize considerably more
languages than other state-of-the-art tools.
Our algorithm and the trained model is provided for academic and personal use.\footnote{https://github.com/tomkocmi/LanideNN}

Since there is no established dataset for the novel
setting of text partitioning by language, we evaluated our model in several common
tasks (monolingual and multilingual language
identification for long and short texts)
which were previously
handled by separate algorithms.
Our approach performs well, improving over the state of the art in
several cases.

A number of things are planned: (1) improving the implementation, especially the
speed of application of a trained model, (2) further extending the set of
covered languages and possibly including more artificial or programming
languages (e.g. JavaScript, PHP) or common formal notations (URLs, hashtags),
(3) evaluating our method on the dataset by
\perscite{king-abney:2013:NAACL-HLT}, possibly
extending this dataset to include more languages,
(4) training and testing the model on noisy texts like Tweets or forum posts,
and
(5) experimenting with other network architectures and
approaches, possibly also training the model on bytes instead of Unicode
characters.

\section{Acknowledgement}
\label{acknowledgement}
This work has received funding from the European Union's Horizon 2020 research and innovation programme under grant agreement no. 645452 (QT21), the grant GAUK 8502/2016, and SVV project number 260 333.

This work has been using language resources developed,
stored and distributed by the LINDAT/\discretionary{}{}{}CLARIN project of the
Ministry of Education, Youth and Sports of the Czech Republic (project LM2015071)

\bibliography{biblio}
\bibliographystyle{eacl2017}

\end{document}